\documentclass{article}
\usepackage{spconf,amsmath,graphicx}

\usepackage{booktabs}
\usepackage{multirow}
\usepackage{amssymb}
\usepackage{enumitem}
\usepackage{diagbox}

\title{Enhancing Domain Generalization in 3D Human Pose Estimation through Controllable Generative Augmentation}
\name{Xinhao Hu, Yiyi Zhang, Liqing Zhang, Jianfu Zhang}
\address{School of Computer Science, Shanghai Jiao Tong University}

\begin{document}
%
\maketitle
\begin{abstract}

Pedestrian motion, due to its causal nature, is strongly influenced by domain gaps arising from discrepancies between training and testing data distributions.
Focusing on 3D human pose estimation, this work presents a controllable human pose generation framework that synthesizes diverse video data by systematically varying poses, backgrounds, and camera viewpoints. This generative augmentation enriches training datasets, enhances model generalization, and alleviates the limitations of existing methods in handling domain discrepancies.
By leveraging both indoor/real-world and outdoor/virtual datasets, we perform cross-domain data fusion and controllable video generation to construct enriched training data, tailored to realistic deployment settings.
Extensive experiments show that the augmented datasets significantly improve model performance on unseen scenarios and datasets, validating the effectiveness of the proposed approach.
\end{abstract}

\section{Introduction}

Pedestrian motion reconstruction and monocular 3D human pose estimation (3D-HPE) are fundamental to autonomous driving, AR/VR, HCI, and robotics, as they enable understanding of human–vehicle interactions and support downstream reasoning. 
Despite notable advances, real-world deployment remains challenging due to  limited multi-view coverage, the high cost of collecting rare events, and substantial domain shifts between training and test data (\textit{e.g.}, indoor vs. outdoor, studio vs. in-the-wild) \cite{saini2019markerless,kim2019pedx}. 
These distributional discrepancies—commonly referred to as the \emph{domain gap}—lead to severe performance degradation when models are applied to unseen domains \cite{wang2018deep,9238468}.

A common remedy is to enlarge training data through augmentation or by merging multiple datasets. Standard 3D motion datasets, such as Human3.6M~\cite{6682899}, MPI-INF-3DHP~\cite{8374605}, 3DPW~\cite{von2018recovering}, and PMR~\cite{wang2025pedestrian}, are each collected under specific capture conditions, which inherently introduce domain differences. Recent domain generalization (DG) efforts for 3D-HPE often operate at the pose level, augmenting or synthesizing joint configurations to increase variability~\cite{zhang2023learning,peng2024dual}. While partly effective, such pose-only augmentation overlooks appearance, background, and camera motion in the video modality, which constitutes a major source of domain shift in real-world applications.

\begin{figure*}[htbp]
    \centering
    \includegraphics[width=1\textwidth]{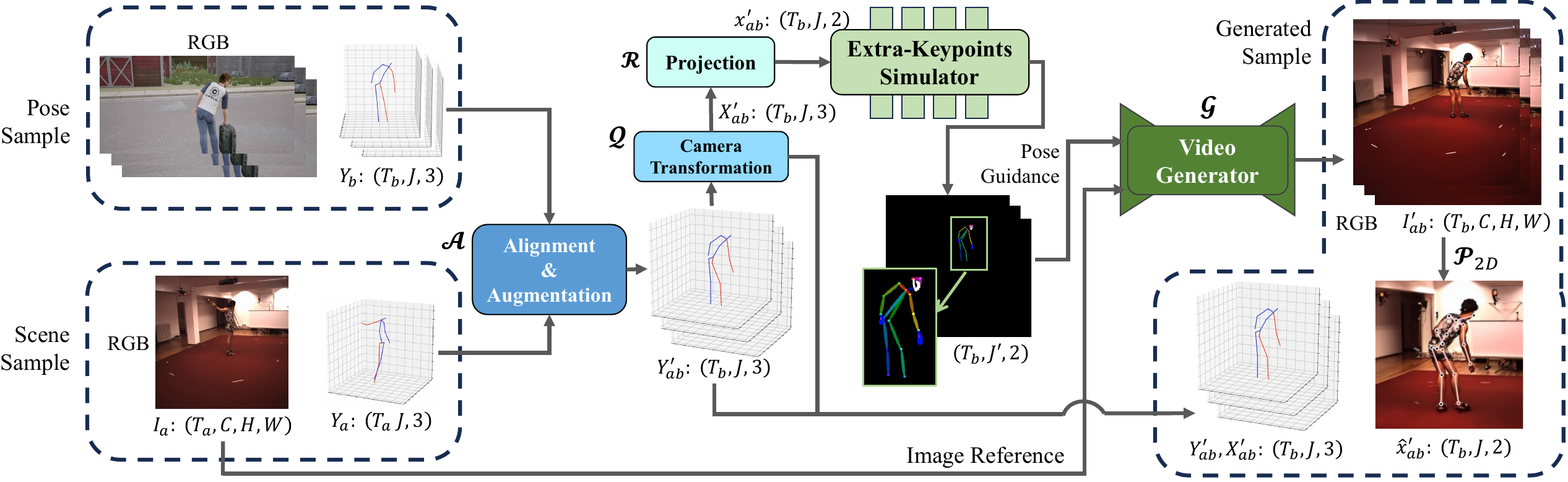}
    \caption{First, the 3D pose sequence from a \textbf{Pose} source sample is aligned to the coordinate frame of a \textbf{Scene} target sample via a simple linear transformation, producing $Y'_{ab}$ (which serves as the 3D ground truth for the new sequence). After projecting to 2D (and simulating any extra keypoints required by the generator), we obtain a pose-guided keypoint sequence $x'_{ab}$. Finally, conditioning on the Scene sample’s RGB frames and the pose guidance, the video generator synthesizes a new video. The output is a paired synthetic sample consisting of the generated RGB sequence and its corresponding 3D pose ground truth.}
    \label{fig:pipeline}
\end{figure*}

Meanwhile, controllable image and video generative models, particularly diffusion-based approaches, have advanced significantly, enabling the synthesis of consistent, temporally coherent videos conditioned on pose, reference appearance, and camera parameters~\cite{ho2020denoising,xu2024magicanimate,hu2023animateanyone}. These models allow fine-grained control over pose, background, and viewpoint, and can thus produce large-scale, automatically annotated RGB sequences without manual collection. While prior work has applied generative models for image-level domain adaptation~\cite{murez2018image,chen2019crdoco}, video-centric generative methods for domain generalization remain largely unexplored.

This work addresses the domain gap by synthesizing large-scale \emph{video} data to expand training domains for 3D-HPE. Nevertheless, using generated videos for 3D-HPE is non-trivial: lifting models are sensitive to the train--test mismatch
between clean projected 2D keypoints and noisy detector outputs, and generative artifacts may further degrade 2D detections.
Motivated by this, we explicitly target the realistic HPE--HPE regime and combine appearance- and camera-aware
video synthesis with a lightweight 2D-consistency filtering step, yielding high-quality synthetic data as an \emph{offline}
augmentation without inference-time overhead.

Rather than augmenting poses alone, we employ controllable video generation to produce diverse RGB sequences through cross-fusion of scenes, poses, and camera trajectories from multiple sources (Human3.6M~\cite{6682899} and PMR~\cite{wang2025pedestrian} in our experiments). The generated videos expose 3D-HPE models to realistic variations in appearance, background, and viewpoint while preserving accurate 2D/3D pose annotations (obtained via the source motion parameters and camera transformations). We then fine-tune state-of-the-art monocular 3D pose estimators (\textit{e.g.}, KTPFormer~\cite{peng2024ktpformer}, D3DP~\cite{shan2023diffusion}, GLA-GCN~\cite{yu2023gla}) and evaluate their cross-domain generalization on held-out datasets. Compared with prior pose-only DG strategies, our video-based approach (1) captures appearance- and camera-induced domain shifts, (2) provides temporally consistent inputs for video-based estimators, thus enabling training on realistic 2D detections, and (3) scales without manual labeling. We leverage recent advances in controllable video diffusion and appearance encoding to preserve subject identity and ensure temporal coherence~\cite{xu2024magicanimate,hu2023animateanyone}. 

Our contributions are threefold:
\begin{itemize}[leftmargin=*]
    \item We propose a pipeline that constructs large-scale synthetic RGB video datasets for 3D-HPE through controllable cross-fusion of poses, scenes, and camera viewpoints.
    \item We show that training with these controllably generated videos (using detected 2D poses as inputs) effectively mitigates domain gaps and substantially improves cross-domain generalization compared to pose-only augmentation, including in realistic settings where both training and testing use detector outputs. 
    \item Extensive experiments on Human3.6M~\cite{6682899} and PMR~\cite{wang2025pedestrian} show that modern 3D-HPE models (\textit{e.g.}, KTPFormer~\cite{peng2024ktpformer}) achieve enhanced robustness on unseen domains.
\end{itemize}

\section{Methodology}

Our goal is to design a practical pipeline that enhances cross-domain generalization in 3D human pose estimation by synthesizing realistic RGB video sequences that integrate scenes, camera parameters, and motions from multiple source datasets, effectively reframing 3D-HPE domain generalization as a video-level augmentation problem. Unlike pose-only augmentation (which ignores scene appearance and camera factors), our approach transplants motions across different scenes and generates large-scale annotated RGB videos with diverse appearances and viewpoints. We then train 3D pose estimators on 2D keypoints detected from these synthetic videos, thereby jointly addressing domain shifts in appearance, background, and viewpoint. 

\subsection{Data and Notation}

We adopt Human3.6M (H36M)~\cite{6682899} and PMR~\cite{wang2025pedestrian} as representative indoor (studio) and mixed-reality outdoor domains. A dataset sample is denoted as $I \in \mathbb{Z}^{N\times T\times C\times H\times W}$ and $Y \in \mathbb{R}^{N\times T\times J\times 3}$, where $N$ is the number of sequences, $T$ is the number of frames per sequence, $C,H,W$ represent the number of channels, height, and width of images, and $J$ is the number of body joints. Ground-truth 3D keypoints $Y$ are transformed into camera space $X$ via $\mathcal{Q}$ and projected to 2D keypoints $x$ via $\mathcal{R}$. Specifically, $\mathcal{Q}$ denotes the transformation into the camera coordinate frame (applying the camera’s rotation and translation), and $\mathcal{R}$ represents the projection from 3D camera coordinates to the 2D image plane (using the camera intrinsics). This explicit modeling of camera pose ensures that our pipeline accounts for viewpoint variations across datasets. A 2D pose detector $\mathcal{P}_{2D}$ maps an image sequence $I$ to detected 2D keypoints $\hat{x}$, and a lifting-based 3D pose estimator $\mathcal{P}_{3D}$ maps $\hat{x}$ to a 3D prediction $\hat{X}$:
\begin{equation}
\mathcal{P}_{2D}(I)=\hat{x},\qquad \mathcal{P}_{3D}:\hat{x}\mapsto\hat{X}.
\end{equation}

\subsection{Generative Augmentation Pipeline}

While existing DG methods mainly augment poses, such strategies neglect scene appearance and camera effects—major sources of domain shift in real deployments. To overcome this limitation, we design a generative augmentation pipeline that explicitly fuses scene, motion, and camera parameters. This enables the synthesis of realistic RGB sequences with preserved 2D/3D annotations, offering richer variability than pose-only augmentation and scaling without manual data collection.

Formally, let $a,b$ index source datasets (\textit{e.g.}, $a=\text{H36M}$, $b=\text{PMR}$) and $m,n$ index samples. A scene sample provides image(s) $I_a^{(m)}$, camera parameters, and root pose $Y_{ao}^{(m)}$; a motion sample provides a full motion sequence $Y_b^{(n)}$ with root $Y_{bo}^{(n)}$. We align the motion to the target scene by translation and a mild linear transform $\mathcal{W}$ (e.g., orientation and ground plane alignment):
\begin{equation}
\mathcal{A}(Y_a^{(m)},Y_b^{(n)})=\mathcal{W}\big(Y_b^{(n)}-Y_{bo}^{(n)}+Y_{ao}^{(m)}\big)=Y_{ab}^{'(l)}.
\end{equation}
The aligned 3D motion $Y_{ab}'$ (which serves as the ground-truth 3D pose for the synthetic sequence) is then projected to 2D keypoints $x_{ab}'=\mathcal{R}(\mathcal{Q}(Y_{ab}'))$ and used as temporal pose guidance for the video generator:
\begin{equation}
\mathcal{G}\big(I_a^{(m,j)},\,\mathcal{R}(\mathcal{Q}(\mathcal{A}(Y_a^{(m)},Y_b^{(n)})))\big)=I_{ab}^{'(l)}\,,
\end{equation}
where $j$ indexes one or more reference frame(s) from the scene acting as appearance anchors. We adopt a controllable video diffusion framework (\textit{e.g.}, AnimateAnyone~\cite{hu2023animateanyone}) to synthesize temporally coherent frames conditioned on the reference appearance and per-frame 2D keypoints.

\begin{figure*}[!t]
    \centering
    \includegraphics[width=\textwidth]{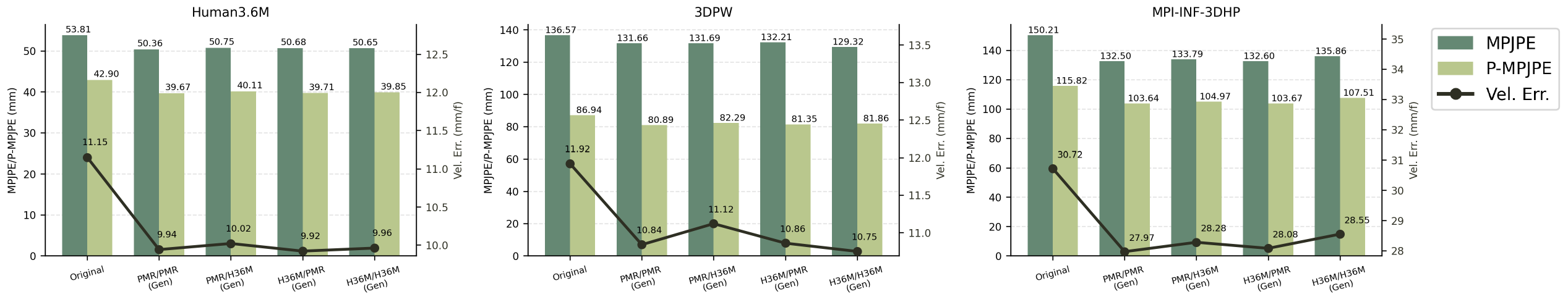}
    \caption{H36M, 3DPW and MPI-INF-3DHP evaluation results of KTPFormer trained on our various augmented datasets (added to the H36M training set).}
    \flushleft\footnotesize{Note: $\mathrm{Gen.}$ denotes generated data. “A/B (Gen.)” indicates generated samples using scene from dataset A and pose from dataset B. \textbf{GT} and \textbf{HPE} denote whether ground-truth or predicted 2D poses (from DWPose) are used as inputs. All results are averaged over sequences. MPJPE and P-MPJPE are in mm, and Vel. Err. denotes the velocity error (mm/frame). Bold values are best (lower is better).}
    \label{fig:main_tab}
\end{figure*}

\subsection{Training Objective}
Applying the above cross-dataset augmentation yields (a) \emph{cross-domain} samples ($a\neq b$) and (b) \emph{in-domain} augmentations ($a=b, m\neq n$). From each generated video $I'_{ab}$ we can obtain either detector-produced 2D poses $\hat{x}'_{ab}=\mathcal{P}_{2D}(I'_{ab})$ or the projected aligned keypoints $x'_{ab}=\mathcal{R}(\mathcal{Q}(Y'_{ab}))$ as inputs to the 2D-to-3D lifting model.

\begin{table}[!t]
\centering
\caption{2D position error between 2D-HPE results of generated samples (under different generator weights and filtering conditions) and the 2D ground truth.}
\resizebox{0.9\columnwidth}{!}{
\begin{tabular}{@{}ccccc@{}}
\toprule
\multicolumn{2}{c}{Generated Dataset} & \multicolumn{2}{c}{\begin{tabular}[c]{@{}c@{}}Original Pretrained \\ Weights\end{tabular}} & \begin{tabular}[c]{@{}c@{}}Fine-tuned on \\ PMR/H36M\end{tabular} \\ \midrule
Scene & Pose & Unfiltered & Filtered & Unfiltered \\ \midrule
PMR & PMR & 87.62 & 17.45 & 20.53 \\
PMR & H36M & 54.52 & 20.50 & 24.25 \\
H36M & PMR & 84.61 & 25.05 & 21.69 \\
H36M & H36M & 64.33 & 27.15 & 24.63 \\ \bottomrule
\end{tabular}
}
\label{tab:filter_comp}
\flushleft\footnotesize{Note: 2D positions are normalized in the camera plane (width $\approx 2000$ pixels, preserving aspect ratio). For filtered samples, only the top 10\% of generated data are retained. \textbf{For the original H36M and PMR test sets, this error is 25.14 and 16.63, respectively}. Thus, our filtered generated data achieves errors close to those of real datasets, ensuring high generation quality.}
\end{table}

Using the generated inputs, we train the 3D pose model with a standard supervised loss:
\begin{equation}
\min_{\theta}\ \mathcal{L}_{\mathcal{P}}\big(\mathcal{P}_{3D,\theta}(\hat{x}'_{ab}),\,X'_{ab}\big)\,,
\end{equation}
where $X'_{ab}$ denotes the camera-space 3D ground truth derived from the aligned motion, and $\hat{x}'_{ab}$ is the detector’s 2D output. This training setup enables the lifter to learn robustness to appearance- and camera-induced variations, complementing pose-only augmentation approaches~\cite{zhang2023learning,peng2024dual,li2023cee,guan2023posegu,huang2022dh}.

\section{Experiments}

\begin{figure*}[!t]
    \centering
    \includegraphics[width=\textwidth]{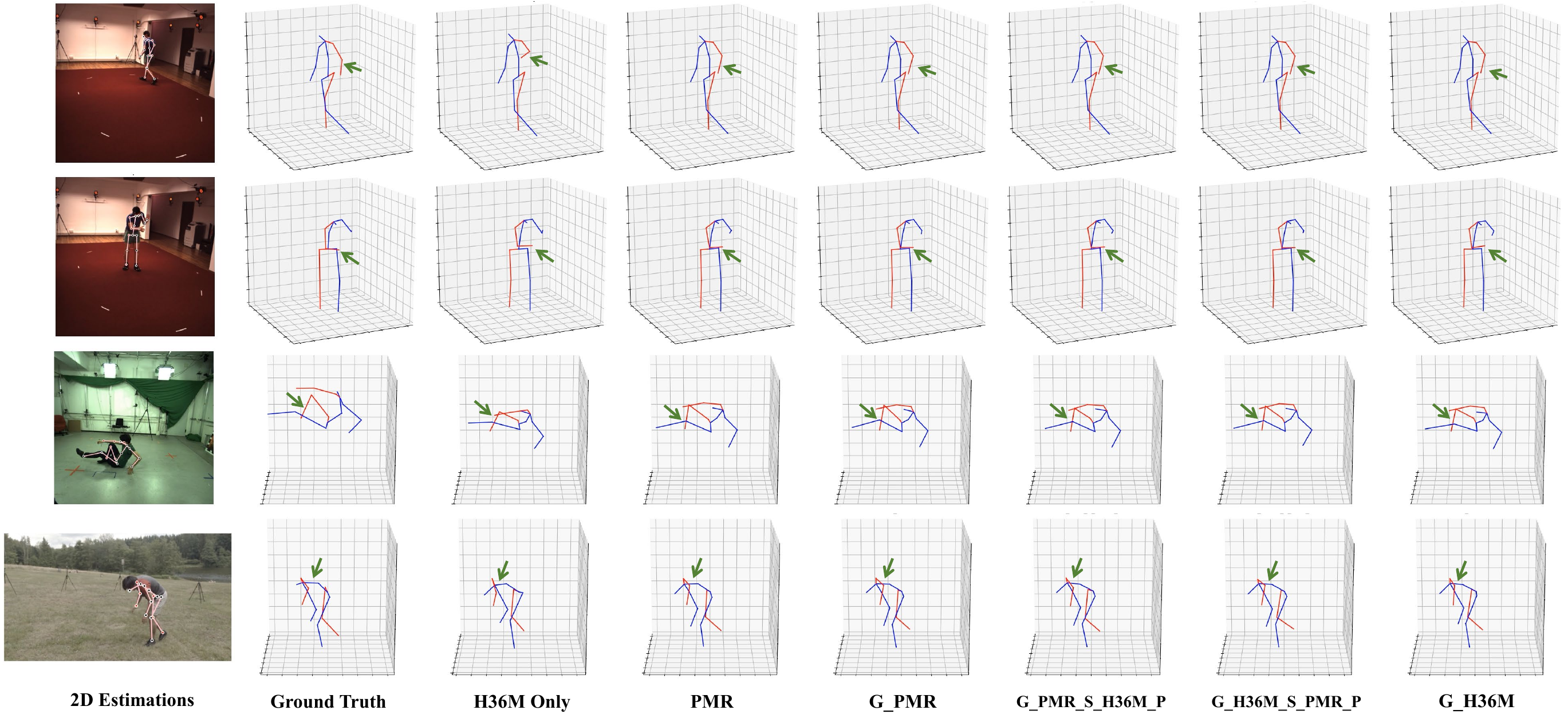}
    \caption{Cross-scenario and cross-dataset performance of KTPFormer on the MPI-INF-3DHP test set after training with our mixed additional datasets combined with H36M (train subjects S1,5,6,7,8). We use 2D-HPE (DWPose) outputs as input. Dataset naming is consistent with Fig.~\ref{fig:main_tab}.}
    \label{fig:h36m_vis}
\end{figure*}

\subsection{Datasets and Implementation Details}
We train lifting-based 3D-HPE models on combinations of H36M~\cite{6682899} and PMR~\cite{wang2025pedestrian} (source domains), and evaluate \emph{cross-scenario} on H36M vs. PMR and \emph{cross-dataset} on MPI-INF-3DHP~\cite{8374605} and 3DPW~\cite{von2018recovering}. Generated RGB videos are produced by cross-fusing scene and pose sources (H36M/PMR) using AnimateAnyone~\cite{hu2023animateanyone}. 2D keypoint detections are obtained with DWPose~\cite{Yang_2023_ICCV}. Since AnimateAnyone requires COCO-WholeBody format~\cite{jin2020wholebodyhumanposeestimation}, we convert the keypoints via a lightweight mapper (simulated hand/face keypoints are dropped). Dataset-specific coordinate handedness is also corrected during alignment. To ensure quality, we filter the generated samples based on the 2D detection error, keeping only the top 10\%. The final augmented corpus (including both H36M and PMR sources) contains about 80,000 frames. All training and evaluation are conducted in PyTorch on an NVIDIA RTX 4090, with KTPFormer~\cite{peng2024ktpformer} as the 3D pose model. Notably, our pipeline does not require fine-tuning the video generator. Generating the full candidate pool took roughly 336 GPU-hours on an RTX 4090 per dataset, but this is a one-off offline process with no impact on inference-time speed. 

During 3D-HPE backbone training and testing, all reported numbers use a single 2D detector (DWPose) to produce
\textbf{HPE} inputs; for \textbf{GT}, 2D keypoints are obtained by projecting 3D ground truth
under the corresponding camera intrinsics/extrinsics. This yields four train/test combinations
(GT--GT, GT--HPE, HPE--GT, HPE--HPE). Importantly, most prior DG methods for 3D lifting
operate only on 2D ground-truth poses during training and therefore can be evaluated fairly only
under the GT--GT/GT--HPE regimes; our pipeline additionally supports the realistic HPE--HPE
setting by generating RGB videos and training the lifter on detector outputs that match the
test-time distribution.

\textbf{Metrics}: 
We use standard MPJPE (mean per-joint position error) and its variants (P-MPJPE after Procrustes, N-MPJPE after normalization):
\begin{equation}
\mathrm{MPJPE}(X,\hat{X})=\frac{1}{NTJ}\sum_{i=1}^{N}\sum_{t=1}^{T}\sum_{j=1}^{J}|X^{(i,t,j)}-\hat{X}^{(i,t,j)}|\,.
\end{equation}
Results are reported as per-sequence averages. We also report Velocity Error (error in first-order temporal differences) and 2D Position Error (MPJPE computed in the image plane).

\subsection{Video Quality and Dataset Variability}
The quality of the generated videos critically affects downstream 2D detections and thus 3D-HPE performance. We measure generation quality by the 2D Position Error between the predicted 2D keypoints and ground-truth 2D keypoints. Table~\ref{tab:filter_comp} shows the error rates under different generator training conditions and filtering strategies. Without filtering, the pre-trained generator’s outputs have high 2D errors ($\sim$54–88 px), far above those on real data (H36M: 25.14 px, PMR: 16.63 px). Applying our 10\% keypoint-error filtering or fine-tuning the generator on H36M/PMR reduces the errors to around 17–27 px (filtered) or 20–24 px (fine-tuned), which is comparable to the real-data baseline. The error distribution for generated samples is heavy-tailed, with most frames around 20–30 px and a long tail of outliers beyond 30 px; filtering effectively removes these outliers. For efficiency, we adopt the filtering strategy by default rather than generator fine-tuning. Hence, all subsequent experiments use the filtered generated dataset. 

\begin{table}[t]
\centering
\caption{MPJPE (mm) on MPI-INF-3DHP test set under different \textbf{Train/Test 2D} settings.
All models are trained on \textbf{H36M + (additional data listed below)}.}
\label{tab:gt_vs_hpe}

\setlength{\tabcolsep}{5.5pt}
\renewcommand{\arraystretch}{1.15}
\small
\setlength{\tabcolsep}{4pt} 
\renewcommand{\arraystretch}{1.1}
\begin{tabular}{lcccc}
\toprule
\multirow{2}{*}{Additional data} &
\multicolumn{4}{c}{Train 2D / Test 2D} \\
\cmidrule(lr){2-5}
& GT--GT & HPE--GT & GT--HPE & HPE--HPE \\
\midrule
PMR                 & 77.07 & \textbf{104.16} & 144.50 & \textbf{127.00} \\
PMR/PMR (Gen.)      & 70.72 & 106.29          & 145.53 & 132.50 \\
PMR/H36M (Gen.)     & \textbf{67.76} & 110.61 & \textbf{143.43} & 133.79 \\
H36M/PMR (Gen.)     & 79.37 & 106.21          & 150.46 & 132.60 \\
H36M/H36M (Gen.)    & 88.54 & 111.39          & 150.98 & 135.86 \\
\bottomrule
\end{tabular}

\vspace{2pt}
\footnotesize
\textbf{Definitions.} GT/HPE indicate whether \textbf{ground-truth 2D keypoints} or \textbf{detector-estimated 2D keypoints} (DWPose) are used as the 3D lifter input.
\textbf{Gen.} denotes generated videos; ``A/B (Gen.)'' means using \textbf{scene from A} and \textbf{pose from B}.
\end{table}

\subsection{Cross-scenario and Cross-dataset Evaluation}

We augment the H36M training set with various generated datasets and train KTPFormer; results are shown in Fig.~\ref{fig:main_tab}. Key observations are: (1) adding our generated data yields small gains on the original H36M test set (improvement of $\approx$3~mm MPJPE under 2D-HPE input) but substantial gains on cross-dataset tests (3DPW: $\approx$4–7~mm; MPI-INF-3DHP: $\approx$15–20~mm), demonstrating improved generalization; (2) different fusion strategies (e.g., PMR/H36M vs. H36M/PMR) produce similar improvements, indicating the robustness of our augmentation pipeline. Qualitative examples (Fig.~\ref{fig:h36m_vis}) show improved local joint accuracy when training includes generated samples. Moreover, models trained with our augmented data exhibit consistently lower velocity errors, indicating smoother motion. For instance, the average velocity error on H36M, 3DPW, and 3DHP improves by 1.19, 1.03, and 2.50~mm/frame, respectively, compared to training without augmentation. This suggests that our synthetic videos help the model predict more temporally coherent pose sequences, while our filtering step effectively culls samples with potential generative artifacts. 

\subsection{Effect of 2D Lifting Source (Training vs. Testing)}

Unlike prior pose-only augmentation methods~\cite{zhang2023learning,peng2024dual,li2023cee,guan2023posegu,huang2022dh}, our approach can be trained and tested using either ground-truth 2D poses (GT) or 2D pose estimates (HPE), yielding four possible combinations (GT–GT, GT–HPE, HPE–GT, HPE–HPE). Table~\ref{tab:gt_vs_hpe} reports KTPFormer’s MPJPE on 3DHP~\cite{8374605} under these different training/testing setups (with various H36M+generated training sets). We observe the following error ordering:
\begin{equation}
(\mathrm{GT,GT})<(\mathrm{HPE,GT})<(\mathrm{HPE,HPE})<(\mathrm{GT,HPE})\,,
\end{equation}
\textit{i.e.}, \textbf{using GT 2D poses for both training and testing yields the lowest error, but in realistic settings where only detector outputs are available, the HPE–HPE setup (matching training and testing distributions) performs best}, while the GT–HPE mismatch gives the worst performance. This finding supports using generated RGB data to provide consistent 2D-HPE inputs in both training and testing.

Notably, prior DG approaches for 3D-HPE that rely on pose-level augmentation cannot train on detector outputs, and thus they face a GT–HPE mismatch at test time. This limitation is inherent: pose-level DG methods augment 2D/3D poses without generating corresponding RGB videos, so they cannot expose the 2D detector (and thus the lifter) to appearance-induced errors during training.

For example, under the same 3DHP evaluation protocol as in Table~\ref{tab:gt_vs_hpe}, 
PoseAug~\cite{zhang2023learning} and the Dual-Augmentor framework~\cite{peng2024dual} 
obtain 142.2~mm and 140.4~mm MPJPE, respectively, in the GT--HPE setting. 
By contrast, our method achieves 132.5--135.9~mm MPJPE in the fully HPE--HPE setting, 
despite relying on detector outputs during both training and testing. 
These results indicate that our generative video augmentation better aligns the 
training distribution with realistic test-time inputs, thereby improving robustness 
to detector-induced noise and domain shift.

\section{Conclusion}

We presented a generative data augmentation approach that expands existing 3D-HPE datasets by cross-fusing samples within and across domains via controllable video generation. Extensive experiments show that the synthesized videos are high-quality, effective for training, and substantially improve model performance on unseen scenarios and datasets. By operating on video data (with varied appearance and camera perspectives), our method complements existing 3D-HPE pose augmentation techniques and can be applied to other vision tasks. Overall, our results highlight the promise of video-level, appearance-aware augmentation for closing the domain gap in 3D-HPE and related fields.

\bibliographystyle{IEEEbib}
\bibliography{refs}

\end{document}